# A Fingerprint Indexing Method Based on Minutia Descriptor and Clustering

Gwang-Il Ri[a, ∗], Chol-Gyun Ri[a] and Su-Rim Ji[b]

[a] *Faculty of Mathematics, Kim Il Sung University, Pyongyang, Democratic People's Republic of Korea*
[b] *Electronic Library, Kim Il Sung University, Pyongyang, Democratic People's Republic of Korea*

## ABSTRACT

In this paper we propose a novel fingerprint indexing approach for speeding up in the fingerprint recognition system. What kind of features are used for indexing and how to employ the extracted features for searching are crucial for the fingerprint indexing. In this paper, we select a minutia descriptor, which has been used to improve the accuracy of the fingerprint matching, as a local feature for indexing and construct a fixed-length feature vector which will be used for searching from the minutia descriptors of the fingerprint image using a clustering. And we propose a fingerprint searching approach that uses the Euclidean distance between two feature vectors as the similarity between two indexing features. Our indexing approach has several benefits. It reduces searching time significantly and is irrespective of the existence of singular points and robust even though the size of the fingerprint image is small or the quality is low. And the constructed indexing vector by this approach is independent of the features which are used for indexing based on the geometrical relations between the minutiae, like one based on the minutiae triplets. Thus, the proposed approach could be combined with other indexing approaches to gain a better indexing performance.

## 1. Introduction

General fingerprint recognition systems involve enrollment and recognition. For the enrollment, one or more fingerprint images are captured from the same finger, features are then extracted from the images and combined with each other to construct a minutia template (an enrolled template) and the template is stored. For the recognition, features extracted from a test image for authentication construct a template, then matching between the test template and the enrolled template is performed. There are two recognition modes – verification and identification. The verification system compares a captured fingerprint image with a selected enrollment template and accepts or rejects the identity claim. Therefore, in the verification system, the extracted template from the test image is compared with only one enrolled template. However, the identification system matches a captured fingerprint image with the entire enrolled templates in the database. Thus, in the identification, the test template should be compared with all the enrolled templates which have already been enrolled in the system.

Many commercial fingerprint recognition systems such as the fingerprint time and attendance system, and the fingerprint access controller system, mainly work in the identification mode. The total number of the fingerprints which have been enrolled in those systems reaches up to hundreds or even thousands of thousand. For those systems where the total number of the enrolled fingerprints is very large, it is important not only to improve the accuracy but also to reduce the response time. If the identification time of a fingerprint recognition system is too long, the system can be hardly used in practice. One simple strategy for the identification is a linear search method where the test template is matched with each enrolled template one-to-one successively. But when the number of enrolled templates is large, the linear search method requires quite a long runtime. In order to speed up the identification, many approaches to reduce the number of the enrolled templates that should be matched with test template have been proposed. One of those approaches narrows down the search range for matching based on the fingerprint classification technique [1]. The fingerprint classification divides all fingerprints into several mutually exclusive classes. In the fast search using the fingerprint classification, the enrolled fingerprints and the test fingerprints are divided into those classes and the test fingerprint is matched only with the enrolled fingerprints of the same class.

Henry's classification is a widely used fingerprint classification. The Henry's classification divides the fingerprints into 5 classes and the fingerprints of only three classes among them dominate almost 90 percent of all the fingerprints. Therefore, a search using this classification can reduce the identification time no more than approximately 1/3. Moreover, if the accuracy of the classification algorithm which automatically classifies the fingerprints is poor, then the recognition accuracy of the overall fingerprint recognition system consequently drops down. In order to overcome limitations of the classification-based search, some approaches were proposed, which construct the features to characterize the fingerprint images with a low computational cost and compare them with each other rather than classify the fingerprints into mutually exclusive classes in order to reduce the number of the enrolled templates to be matched 1:1. Those approaches are called a fingerprint continuous classification or a fingerprint indexing.

In [3, 11, 12], the authors summarized studies on the fingerprint indexing. In [3], approaches for the fingerprint indexing are classified into minutia-based ones, ridge-based ones, transform-based ones, SIFT (Scale Invariant Feature Transform) feature-based ones and so on. SIFT feature consists of keypoints and keypoint descriptors and is extracted using four steps. The first step is the Scale-space extrema detection and the second step is the keypoint localization. The third step is the Orientation assignment and the fourth step is construction of the keypoint descriptor. Typical examples of the minutia-based approaches are the minutiae triplets-based ones. In those approaches, the minutiae triplets are generated from minutiae extracted from fingerprint images and for every minutiae triplets the feature vectors to characterize the triplet are extracted.

The ridge-based approaches extract features for indexing from ridge structures around every minutia. In [3], the authors showed that the minutiae-based approaches and the ridge based approaches are both more effective than the other ones to narrow down the search range through experiments.

In [11], the authors divided the fingerprint indexing approaches into local feature-based ones, global feature-based ones and other feature-based ones. The orientation field and the ridge frequency field and the information of singular points belong to the global features. Since the global feature-based approaches need information of position and direction of the singular points, etc., if the test fingerprint image does not contain the singular points, then the approaches cannot be applied and of course their effectiveness requires the high accuracy of information for singular points.

In [11], the authors concluded that the performance of the local feature-based fingerprint indexing is the best. The minutia information such as the position and direction of a minutia, the features of the ridge around minutia or of image are typical of the local features.

There have been many attempts to improve the effectiveness of the local feature-based indexing. In [9], the authors made the minutia detail for every minutia using the relation between the ridges connected to the minutia. After that, they performed the Delaunay minutiae triangulation, extracted the geometrical features for characterizing every minutiae triplet and combined them with the minutia detail to represent the minutiae triplet. The search is performed by comparing those minutiae triplets. In [5], the authors constructed the substructure from the ridges around every minutia and extracted invariant relations between the ridge substructures as the feature for searching. In [4], the authors extracted the features based on the Minutia Cylinder-Code for every minutia, performed the search using a hashing scheme and showed the effectiveness of their approach. In [14], the authors describe every minutiae triplet as a 9-D feature vector which contains the curvature information of the ridge connected to the minutia. Then they cluster these feature vectors in 9-D space. Finally, the index of cluster where every minutiae triangle belongs to is used as the feature for fingerprint indexing. In [6], for every minutia they formed the binary bit array based on geometrical information of neighbors minutiae around it and performed the search using a hash function. In [1], the authors proposed a scheme which adds the curvature information of the ridge connected to every minutia into the minutiae triplet-based fingerprint indexing and showed that the adding of the curvature information enhances the performance of indexing. In [13], the authors proposed an approach which uses level-1 feature, level-2 feature and performed the search using a hash function and showed that the combining two features enhances the indexing performance more than using separately the features. In [8, 10], the authors represented the fingerprint images by using the feature of the ridge orientation field and the average distance between ridges and used the clustering in the feature space of the ridge orientation field. The search is performed by finding the cluster which the test image corresponds to and matching only with the enrolled template in the cluster.

In [2], the authors proposed a new fingerprint indexing approaches based on orientation fields, fingerprint codes and the minutiae triplets and showed that these approaches were more effective than the linear search. They also proposed a fingerprint indexing approach which combines three features and showed that the combining is better than the individual feature-based one in the performance of indexing.

In [15], the authors proposed the fusion methods of level-1 and level-2 indexing method and showed that the fusion method is more effective than level-1 and level-2 method. The level-1 method uses orientation and frequency fields as feature, and level-2 method uses Minutia Cylinder-Code. The authors evaluated several fusion methods and showed that the hybrid fusion method is best accurate through experiments on six databases.

From the above consideration, we can find the following facts.

(a)    The local features which represent the minutiae triplets or the ridges structure around every minutia are effective local features for indexing.

(b)    Combining the global or local features which are different from the feature that describes the minutiae triplet can improve the performance of indexing. Especially, combining the local features which describes every minutia is the best approach to improve the performance of indexing.

(c)    Typical approaches for the search in indexing are the hash function-based ones and the clustering-based ones.

In this paper we propose a new approach that uses a minutia descriptor which describes the characteristic of every minutia as the local features for indexing and constructs the feature vector of a fixed length for the search by clustering.

Constructing the features and searching method are vital to the local feature-based fingerprint indexing.

In Section 2, we explain what should be considered for selecting both of these (constructing the features and searching method) in the local feature-based indexing. In Section 3, we describe how to construct the feature vector for indexing by the proposed approach in this paper. In Section 4, we show the experimental results. In Section 5, we show conclusions.

## 2. The selection of local features and searching methods for the fingerprint indexing

The purpose of the fingerprint indexing is to reduce the identification time without the loss of accuracy of the fingerprint recognition system. Since a search with the fingerprint indexing is performed by comparison the indexing features, we can regard the fingerprint indexing as a kind of matching process. Thus, we can think that the fingerprint recognition with the fingerprint indexing uses two stages of matching for a speedup in the identification. The first stage of matching is the fingerprint indexing and it has a low computational cost, while the second stage is the original 1:1 matching and it has a high accuracy, however, the matching time is long. Generally, 1:1 matching also consists of several stages for a speedup. Going from the previous stage to the next stage makes the computational cost increased and the accuracy higher. If the similarity of the previous stage is lower than a threshold, then the next stages of matching will be canceled. Recently, it has been showed that combining different indexing methods enhances the indexing performance. Therefore, the indexing can also involve several stages, each of which is also indexing.

The fingerprint identification with indexing includes several stages of the matching and the earlier stages of matching can be regarded just as comparison between indexing features. Figure 1 shows several stages of matching for the recognition system with the fingerprint indexing.

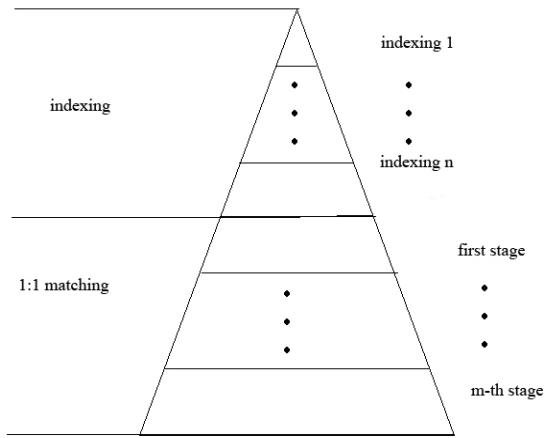

**Fig. 1.** Several stages of matching in the fingerprint recognition.

Taking the above consideration into account, the fingerprint indexing which is the early stage among several stages of the matching should have the following properties.

i. *The stability*
   The fingerprint indexing should retain the enrolled template of the same fingerprint as the fingerprint of the test image as much as possible. In other words, the indexing features of the same fingerprints from different images should be similar with each other, irrespective of the size and the quality of the images.

ii. *The independence*
   If the features which are used in the different stage involves the independent information, then the identification speed of the system will be faster. For example, if a fingerprint recognition system has two stages of indexing and the features which are used in different stage of indexing are independent of each other, then the indexing performance will be improved more than when they have a high correlation.

iii. *The distinguishing ability*
   The indexing should have a distinguishing ability between the different fingerprints. That is, the indexing features of the same fingerprints from the different images should be similar with each other and the indexing features of the different fingerprints should be different from each other.

iv. *The fast matching*
   The matching for indexing features should be simple and fast.

The indexing feature based on the minutiae triplets and the indexing feature based on the Minutia Cylinder-Code have a high distinguishing ability. They employ the geometrical relations between the minutiae as features. In general, these features are also employed in 1:1 matching later. Moreover, since they use the mutual relations of the minutiae, they are largely influenced by adding of the false minutia or removal of the true minutia, further, the performance of them relies heavily on the size and the quality of the fingerprint images. The approaches which extract the indexing features from the ridge structure around minutia need the ridges to be accurately extracted from fingerprint images. These approaches are also heavily influenced by the quality of fingerprint images.

The purpose of this paper is to construct the stable indexing features with high distinguishing ability without use of the geometrical relations between the minutiae. If we employ these features in conjunction with other indexing which uses the mutual relations between the minutiae, the identification speed of a system will be faster. In this paper, for every minutia, we construct a minutia descriptor which robustly represents characteristics of the image around the minutia and adopt it as the indexing features.

The typical approaches for the comparison of the local feature-based indexing features include the hash-based one and the cluster-based one. Both of them divide the local features for indexing into mutual exclusive classes and perform the search using indexes of the classes where the local features of fingerprint image belong to. Here, the operation which assigns local features to a class is not robust. Especially, a small number of local features makes things worse. Some clustering-based search approaches assign every local feature to several classes rather than assign to only one class. These approaches improve the effectiveness of searching in a certain grade, but they have the computational cost increased and still remain the robustness problem unsolved.

In this paper, we construct a feature vector of a fixed length for the search by using the clustering in order to solve these problems. The proposed approach uses the clustering of the local features, but it is different from the previous papers in the manner of exploiting the results of the clustering.

**3. Construction of an indexing feature vector**

In this section, we describe a minutia descriptor that is a local feature for the fingerprint indexing and the method for constructing a feature vector of a fixed length for the search.

*3.1. The minutia descriptor*

In our previous study [16], we constructed a minutia descriptor which describes the characteristics of the image around the minutia for every minutia of the fingerprint image in order to improve the accuracy of the fingerprint recognition system. This minutia descriptor is robust since it only uses positional and directional information of the minutia and the image around the minutia. And the distinguishing ability of the descriptor is improved by using the Gabor wavelet and Linear Discriminant Analysis. The proposed minutia descriptor is a vector of a fixed length. Below we explain the procedure for constructing the minutia descriptor.

- Extracting minutiae

Extracting minutiae is performed by the algorithm of paper [7].

- Enhancement of the fingerprint image

For the fingerprint image, first, DOG (Difference of Gaussian) filtering is performed and next, local mean variance normalization is done. DOG filtering is performed by first performing two Gaussian filtering with different size, and then calculating difference of them. DOG filtering is one kind of band pass filtering. Figure 2 shows the result of an image enhancement. In the Figure the left is the image before the enhancement and the right is the image after the enhancement.

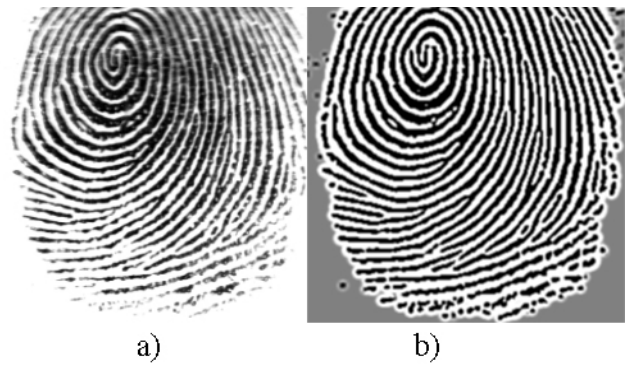

**Fig. 2.** Result of an image enhancement. a) original image b) enhancement result of image a)

- **Calculation of Gabor wavelet coefficients**

For every minutia, we take 9 sampling points adopting the minutia as the center and the direction of the minutia as the reference direction as shown in Figure 3.

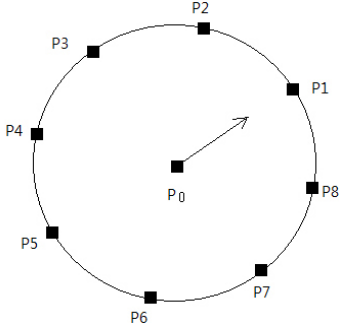

**Fig. 3.** Sampling points around a minutia.

In the Figure, $P_0$ is the central minutia and the arrow is the direction of the minutia. The Gabor wavelet coefficients are calculated from the enhanced image. For all the 9 sampling points, the absolute values of the Gabor wavelet coefficients of 5 frequencies and 8 directions are calculated. Since 40 absolute values are obtained for every sampling point, we concatenate them to get a 360-dimention feature vector of Gabor wavelet.

- **Construction of the Linear Discriminant Analysis transform**

In order to perform the Linear Discriminant Analysis, we built a database for training by an optical fingerprint sensor. The database consists of images of 500 fingers and contains 10 images for every finger. We extract minutiae from every image of the database using the algorithm of paper [7] and calculate a 360-D feature vector of Gabor wavelet for every minutia. Next, we perform the Principal Component Analysis in the 360-D space taking the set of all the feature vectors as the training data set and get a Principal Component Analysis transform matrix which converts a 360-D vector to a 30-D vector. We get 30-D feature vectors from all the 360-D feature vectors of the Gabor wavelet by Principal Component Analysis. And we perform Linear Discriminant Analysis in the 30-D space. Here, every class is formed from the 30-D feature vectors which are calculated for same minutia of one finger. Therefore, the number of sample vectors in every class is 10 at the most. The minutiae from different fingers correspond to the different classes. The judgment whether the minutiae are same or not could be determined manually or automatically by matching. The Linear Discriminant Analysis is performed in the 30-D feature space to get a linear transform from 30-D to 25-D feature vector. Finally, we get a 360×25 linear transform matrix by multiplying the principal component transform (360×30) and the linear discriminant transform (30×25)

- **Performing a Linear Discriminant Analysis to the Gabor wavelet feature vectors**

For every minutia, the above constructed linear transform is applied for the 360-D Gabor wavelet feature vector to get a 25-D vector. This 25-D vector is just the descriptor of the minutia.

*3.2 Construction of the feature vector of a fixed length for the search*

In the previous papers which proposed the clustering for the search in the fingerprint indexing, the authors performed clustering in the local feature space and represented every fingerprint image using the indexes of the clusters where the local features belong to. In this paper, we perform the clustering in the local feature space and form a stable feature vector of a fixed length for the search using the clustering result and the local features. Below, we explain how to construct the feature vector for the search.

- **Clustering in the local feature space**

In order to perform the clustering of the local feature vectors we reuse the training database that we used before in subsection 3.1. First, minutiae from all fingerprint images in the database are extracted. For every extracted minutia the above mentioned 25-D minutia descriptor is calculated. Then, all the calculated minutia descriptors are collected to get a training set for the clustering. 200 central vectors of the clusters are obtained by performing the k-mean algorithm on the training set. Every central vector is a 25-D vector.

- **Feature vector for the search**

Let $\{C_k \mid k=1,\cdots,200\}$ denote the 200 central vector of 25-D. Let $N$ and $\{V_i \mid i=1,\cdots,N\}$ denote the number of the extracted minutiae from the fingerprint image and the minutia descriptor vectors corresponding to the minutiae, respectively.

First, vectors $M^i = \{M_k^i \mid k=1,\cdots,200\} \; i=1,\cdots,N$ are calculated. These vectors represent the degrees that every minutia descriptor vector belongs to every cluster. The equations are as follows.

- $d_k^i = \|C_k - V_i\|^2, k=1,\cdots,200$ (1)

- $md^i = \min_k d_k^i$ (2)

- $d_k^i = d_k^i - md^i, k=1,\cdots,200$ (3)

- $d_k^i = \exp(-d_k^i), k=1,\cdots,200$ (4)

- $S^i = \sum_{k=1}^{200} d_k^i$ (5)

- $M_k^i = d_k^i / S^i$ (6)

Using the calculated vector set $\{M^i \mid i=1,\cdots,N\}$, a 200-D vector $F = \{F_k \mid k=1,\cdots,200\}$ for the search is computed as follows.

- $SM = \sum_{i=1}^{N} M^i$ (7)

Here, $SM = \{SM_k \mid k=1,\cdots,200\}$ is a 200-D vector.

- $s = \dfrac{\sum_{k=1}^{200} SM_k}{200}$ (8)

- $SM_k = SM_k - s, \; k=1,\cdots,200$ (9)

- $ss = \sqrt{\dfrac{\sum_{k=1}^{200} SM_k^2}{N}}$ (10)

- $F_k = SM_k / ss, \; k=1,\cdots,200$ (11)

The obtained vector $F$ is a 200-D vector formed from every fingerprint image for the search.

*3.3 Searching method*

- Construction of a search feature vector for the enrolled template.

If one fingerprint image is used for every finger enrollment, then the search feature vector is constructed by the above mentioned method using the minutia descriptors of every minutia in the enrolled template. In the case where multiple fingerprint images are used for every finger enrollment, we propose the method that first constructs feature templates from fingerprint images, then combines the templates to one super-template and finally makes the search feature vector using this super-template. The procedure for constructing the super-template is as follows:

    a) Select one template. Set this template as initial super-template.

    b) For every other template,

        Perform matching it with current super-template, and minutiae correspondence.

        For correspondent minutiae pairs, minutia information (position, direction) and minutia descriptor is modified by using averaging.

        For non-correspondent minutia of the template, add it to super-template (with its minutia descriptor).

- Search

    For the search, a constant $0 < \Pr \leq 1$ is fixed in advance. This constant represents the percentage of the enrolled templates which should be 1:1 matched. The minutia-based template is formed for the test fingerprint image and the feature vector is obtained for the search. Then, the Euclid distance between this feature vector and the search feature vector of the enrolled template is calculated. Next, the enrolled templates are sorted according to their distances in the increasing order. If the number of the enrolled templates in the fingerprint recognition system at present is $N$, then only first $\lceil N \times \Pr \rceil$ templates is 1:1 matched.

**4. Experiments**

In experiments, we test the effectiveness of the proposed fingerprint indexing on six databases.

These databases are FVC 2000 DB2 A, FVC 2000 DB3 A, FVC 2002 DB1 A, FVC 2004 DB1 A, FVC 2004 DB2 A, FVC 2006 DB2 A. First five databases are formed with the fingerprint images of 100 fingers and it includes 8 fingerprint images for every finger. FVC 2006 DB2 A is formed with the fingerprint images of 140 fingers and it includes 12 fingerprint images for every finger. For every database, we convert the resolution of every image to 500dpi and then crop the images of size 320×320 around the center of the fingerprint region.

The sample images of the databases are shown in Figure 4.

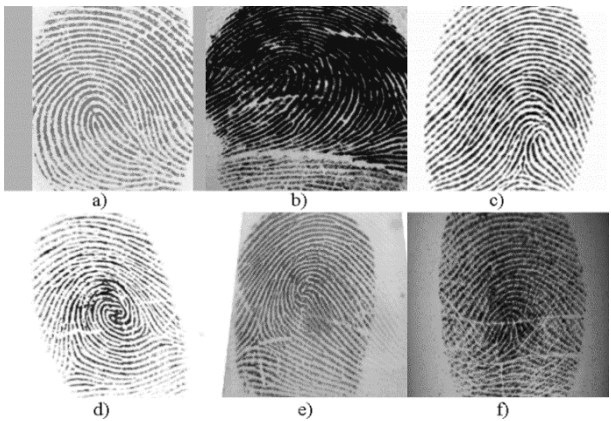

**Fig. 4.** Sample images of six databases, a) FVC2000 DB2 b) FVC2000 DB3, c) FVC2002 DB1 d) FVC2004 DB1 e) FVC2004 DB2 f) FVC2006 DB2

In the experiments, we compare the indexing performance for two fingerprint recognition systems with other methods. In the first system, the enrollment is performed using the first image and the search is done using images from second to the last for every finger. In second system, the enrollment is performed using the first three images and the search is done using images from 4-th to the last for every finger. Here, for the enrollment, the three templates are merged to get a super-template as we mentioned before.

The indexing performance is evaluated by Penetration Rate vs Error Rate curve.

This curve shows error rate at every fixed penetration rate. The penetration rate is the portion that the fingerprint recognition system has to perform matching after indexing (sort). The error rate is the portion of fingerprint images that not in the search range (determined by penetration rate) after indexing.

For evaluating the search time, we construct the other fingerprint image database. This database is consist of the fingerprint images of 2000 fingers, and contains 3~100 images for every finger. The resolution of the image is 500dpi and the size is 320×320. At this time, we enroll every finger using three images. The table 1 compares the search time of proposed method with the previous methods.

The average time for calculating minutiae descriptors and adding the search feature vector for one template is about 4ms (on Computer Intel Core i7 @ 2.7Ghz).

From above experiment results, we conclude as follows:

- The enrollment using three images from same finger remarkably improves the indexing accuracy than enrollment using one image.

- The indexing accuracy is comparable with previous methods except MCC based method by Cappelli et al.

- The search time of proposed method is very short.

The proposed method can be used alone as indexing method in good condition, but it can be combined with other indexing methods for future improvement.

**5. Conclusion**

In this paper we proposed a robust fingerprint indexing approach using a minutia descriptor and clustering. We employ the minutia descriptor which is constructed using the positional and directional information of the minutia and the neighbor patch around the minutia, as a local feature information. And we proposed an approach which forms a robust vector of a fixed length for the search from the minutia descriptor using the clustering. We demonstrated the effectiveness of the proposed fingerprint indexing approach on six databases. The proposed indexing approach does not employ geometrical relations between minutiae. Thus, if the approach is combined with the other fast matching approaches which employ geometrical relations between minutiae, it should be possible to get a dramatic speedup of the identification. In the future, we will study a method how to combine the proposed approach here with the other approaches.

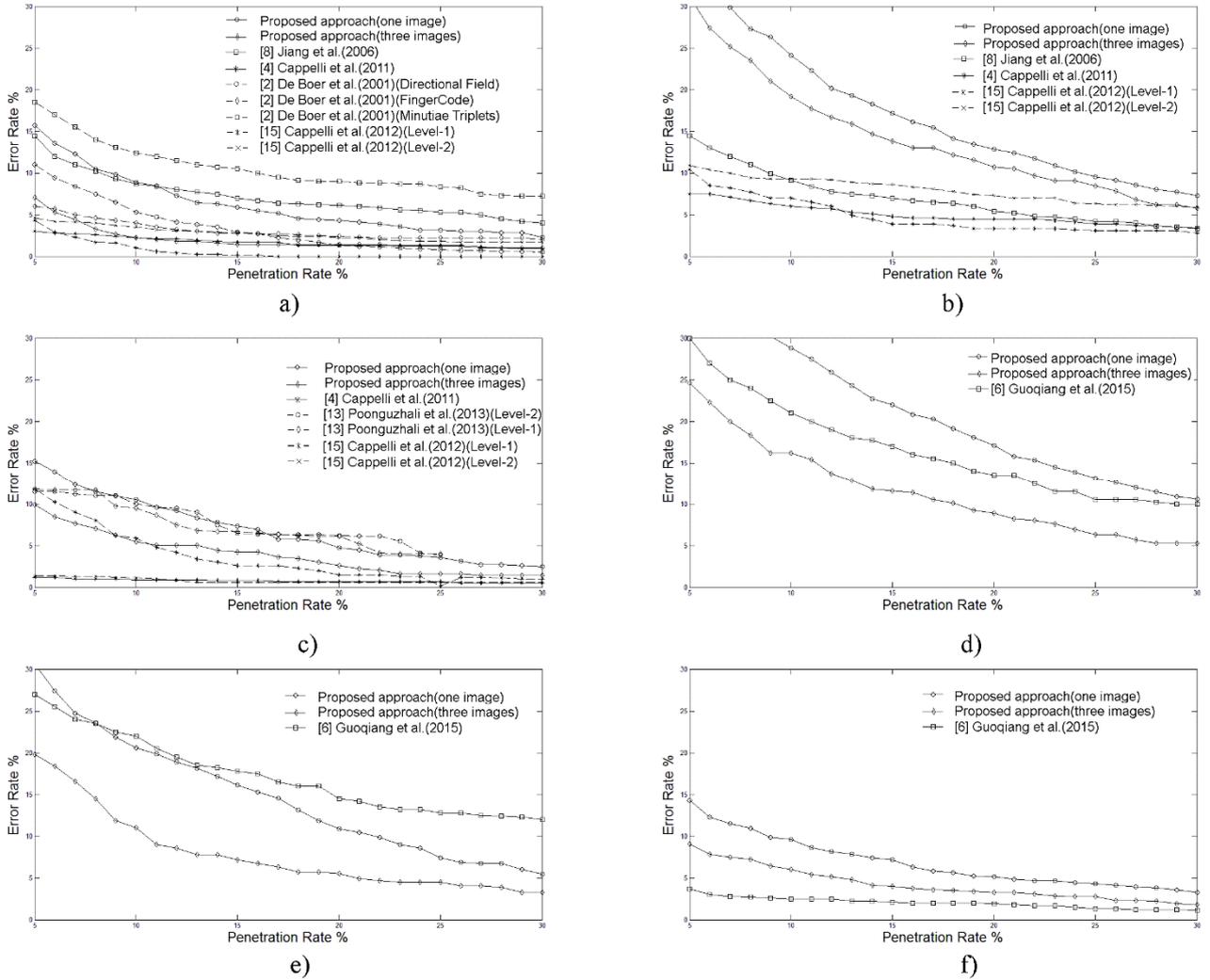

**Fig. 5.** Indexing performance a) FVC 2000 DB2 A b) FVC 2000 DB3 A c) FVC 2002 DB1 A d) FVC 2004 DB1 A e) FVC 2004 DB2 A f) FVC 2006 DB2 A

**Table 1. Search Time**

| Method | Search Time (ms) | Hardware | Database | Number of fingers |
|---|---|---|---|---|
| [8] Jiang et al.(2006) | 67 | Intel Pentium 4 @ 2.26 Ghz | NIST DB4 | 2000 |
| [15] Cappelli et al. (2012) Level-1 | 1.6 | Intel Core 2 Quad @ 2.66 Ghz | NIST DB4 | 2000 |
| [15] Cappelli et al. (2012) Level-2 | 14 | Intel Core 2 Quad @ 2.66 Ghz | NIST DB4 | 2000 |
| [15] Cappelli et al. (2012) Fusion | 16 | Intel Core 2 Quad @ 2.66 Ghz | NIST DB4 | 2000 |
| Proposed method | 0.3 | Intel Core i7 @ 2.7Ghz | Our own | 2000 |